\documentclass{article}
\usepackage[table, dvipsnames]{xcolor}

\usepackage{microtype}
\usepackage{graphicx}
\usepackage{booktabs} %

\usepackage{hyperref}

\usepackage[accepted]{codeml}

\usepackage{amsmath}
\usepackage{amssymb}
\usepackage{mathtools}
\usepackage{amsthm}
\usepackage{caption}
\usepackage{subcaption}
\usepackage{multirow}
\usepackage{graphicx}
\usepackage{enumitem}
\usepackage{lipsum}

\usepackage{algorithm}
\usepackage{algorithmic}
\newcommand{\name}[1]{Spatial Reasoners}

\definecolor{boxyellow}{RGB}{251, 246, 192}

\newcommand*{\img}[1]{%
  \raisebox{-.22\baselineskip}{%
    \includegraphics[
      height=\baselineskip,
      keepaspectratio,
    ]{#1}%
  }%
}

\newcommand{\namelogo}{\img{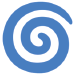}\,\textbf{\name}}

\usepackage[capitalize,noabbrev]{cleveref}

\theoremstyle{plain}

\theoremstyle{definition}

\theoremstyle{remark}

\usepackage[textsize=tiny]{todonotes}

\icmltitlerunning{\name{} for Continuous Variables in Any Domain}

\begin{document}

\twocolumn[
\icmltitle{\namelogo{} for Continuous Variables in Any Domain}

\begin{icmlauthorlist}
\icmlauthor{Bart Pogodzinski}{yyy}
\icmlauthor{Christopher Wewer}{yyy}
\icmlauthor{Bernt Schiele}{yyy}
\icmlauthor{Jan Eric Lenssen}{yyy}
\end{icmlauthorlist}

\icmlaffiliation{yyy}{Max Planck Institute for Informatics, Saarland Informatics Campus, Germany}

\icmlcorrespondingauthor{Bart Pogodzinski}{bpogodzi@mpi-inf.mpg.de}
\icmlcorrespondingauthor{Christopher Wewer}{cwewer@mpi-inf.mpg.de}
\icmlcorrespondingauthor{Bernt Schiele}{schiele@mpi-inf.mpg.de}
\icmlcorrespondingauthor{Jan Eric Lenssen}{jlenssen@mpi-inf.mpg.de}

\icmlkeywords{Machine Learning, ICML, Generative Models, Spatial Reasoning, Software framework, SRM}

\vskip 0.3in
]

\printAffiliationsAndNotice{}

\begin{abstract}
We present \namelogo{}, a software framework to perform spatial reasoning over continuous variables with generative denoising models. Denoising generative models have become the de-facto standard for image generation, due to their effectiveness in sampling from complex, high-dimensional distributions. Recently, they have started being explored in the context of reasoning over multiple continuous variables. Providing infrastructure for generative reasoning with such models requires a high effort, due to a wide range of different denoising formulations, samplers, and inference strategies. Our presented framework aims to facilitate research in this area, providing easy-to-use interfaces to control variable mapping from arbitrary data domains, generative model paradigms, and inference strategies. \name{} are openly available online\footnote[2]{spatialreasoners.github.io}.

\vspace{-0.3cm}
\end{abstract}
    
\section{Introduction}
\label{sec:introduction}
Denoising generative models, such as DDPM~\cite{ddpm}, DDIM~\cite{ddim}, Flow Matching~\cite{lipman2023flow}, or Rectified Flow~\cite{liu2022rectifiedflow} have achieved unmatched levels of generation quality, and the research work in this field only continues to accelerate. 
Typically, these models learn to approximate a conditional data distribution $p(x \mid c)$ and learn to sample from it, where $x$ represents a variable like images and $c$ can be text or other conditioning signals.

In the recent year, the trend evolved further and interest grew in diffusion models that allow sampling over multiple variables, where each has its own noise level~\cite{chen2024diffusionforcingnexttokenprediction, ruhe24rolling, wewer25srm}. This scheme allows a wide range of sampling techniques, such as auto-regressive generation (with planned order), generation with infinite horizon, and overlapping generation, essentially turning denoising models into an engine for general probabilistic inference. Spatial Reasoning Models (SRMs)~\cite{wewer25srm} formalized this framework into a general variant of such models that, given some partitioning of the data format into variables $\{x_1,...,x_n\}$, e.g. image patches, video frames, skeleton joint positions, language tokens, etc., allows sequential conditional inference across these variables by decomposition using the chain-rule of probability:
\begin{equation}
p(x_1,..,x_n) = \prod^n_{i=1} p(x_{\pi(i)} | \{x_{\pi(j)}\}^{n}_{j=i+1})
\end{equation}
As shown, optimizing the specific inference strategy, such as order and amount of sequentialization, can significantly reduce hallucinations in the generations~\cite{wewer25srm}. 

\vspace{-0.1cm}

Currently there are many different diffusion formulations, noise schedules, samplers, and inference variants. Thorough analysis and adaption to other data domains requires large-scale ablations and significant implementation effort. 
It is common that in the research field rapid development is prioritized over good separation of concerns, modularity and readability. This allows quickly testing ideas, but makes it harder to build on top of them. We believe an intuitive, modular and expandable framework and project template would therefore immensely help to further develop the paradigm of reasoning with denoising generative models. 

\vspace{-0.1cm}
In this work, we present \namelogo{}, a software framework for performing spatial reasoning over sets of continuous random variables via multi-noise-level denoising generative models. We hope it will facilitate solving generative tasks in a wide range of new domains that go beyond image representations. 
\begin{figure*}[t]
\centering
\includegraphics[width=\textwidth]{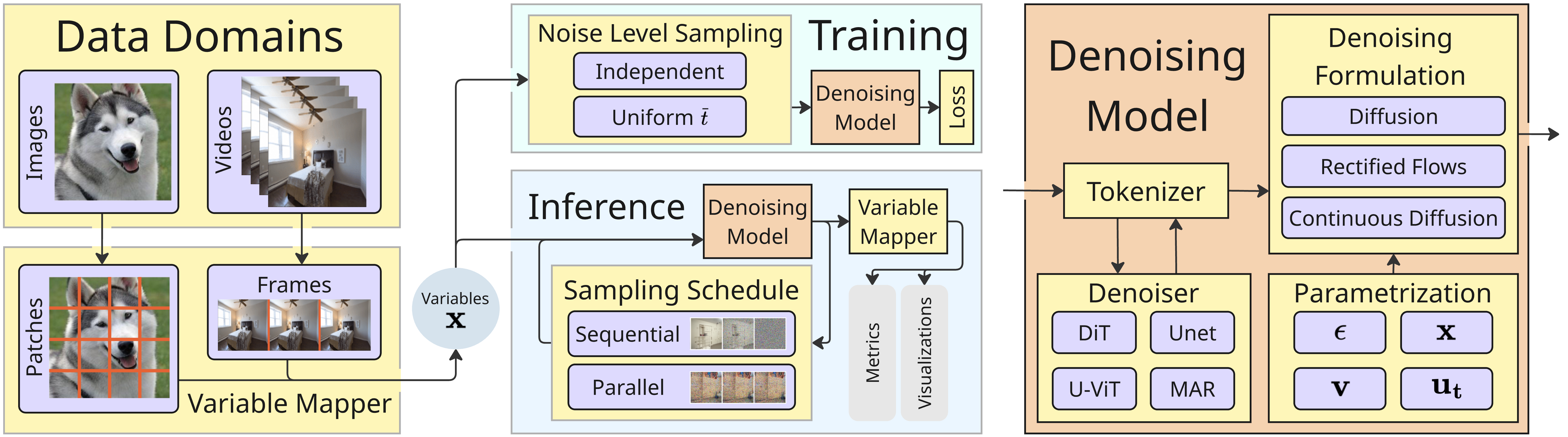}
\vspace{-0.3cm}
\caption{\textbf{Overview of \name{}}. The variable mapper transforms different input modalities, such as images, videos or others, into variables, which are then used for training and inference. Yellow \textcolor{boxyellow}{$\blacksquare$} building blocks are exchangeable and can be extended with custom functionality. On the right, we zoom into individual aspects of the denoising model.
\vspace{-0.4cm}}
\label{fig:pipeline}
\end{figure*}
\name{} expose the following degrees of freedom in an easy-to-use interface: 
\vspace{-0.4cm}
\begin{itemize}
\setlength\itemsep{0em}
    \item The choice of the input domain by providing a generic mapper interface that transforms arbitrary data domains into sets of variables to reason over.
    \item Explicit control over training and inference schedules, e.g., order and amount of sequentialization, individual noise levels, and the denoising formulation.
    \item A range of denoiser architectures, such as UNet~\cite{ldm}, DiT~\cite{dit}, LightningDiT~\cite{lightningdit}, U-ViT~\cite{simple_diffusion, song2025historyguidedvideodiffusion}, MAR~\cite{li2024autoregressive}, xAR~\cite{ren2025nexttokennextxpredictionautoregressive}, and AEs~\cite{dcae, ldm, lightningdit} for latent modelling, to be used depending on modality and task.
\end{itemize}

\section{Related Work}
\label{sec:related_work}

Applying denoising generative models with individual noise levels per variable has been a recent trend in several works. Rolling Diffusion \cite{ruhe24rolling} and Diffusion Forcing \cite{chen2024diffusionforcingnexttokenprediction} have proposed video diffusion models that are trained to denoise different noise levels per frame, allowing to generate long sequences in a rolling window. 
History-guided diffusion~\cite{song2025historyguidedvideodiffusion} recently expanded on the video generation with the Diffusion Forcing scheme, by exploring the impact of classifier-free guidance with respect to clean frames.  MAR~\cite{li2024autoregressive} and xAR~\cite{ren2025nexttokennextxpredictionautoregressive} present auto-regressive diffusion on images that denoise a single or $k$ variables at the same time. Spatial Reasoning Models (SRMs)~\cite{wewer25srm} present a general framework for such strategies on sets of variables.
Another line of work involves denoising models applied across multiple modalities. UniDiffuser \cite{bao2022one} performs diffusion jointly on text and images, by independently sampling the noise level for each modality. The model can be used to conditionally generate images from given text, vise versa, or joint generation.
\name{} unify these paradigms into a single framework, allowing to mix and explore individual choices, such as inference schedules, architectures, and denoising models. 

\vspace{-0.4cm}
\paragraph{Other packages.} Related software frameworks for denoising generative models have been very successful in recent years. Examples include HuggingFace Diffusers~\cite{von-platen-etal-2022-diffusers}, a framework for diffusion models for image generation, which supports most of the research in this domain. Another widely used toolkit is the \texttt{denoising-diffusion-pytorch} repository~\cite{denoising-diffusion-pytorch}. None of the existing frameworks explicitly supports generation and reasoning across multiple variables.

\section{Spatial Reasoning with \name{}}
\label{sec:method}
In this section, we first introduce the core reasoning framework of \name{} in Sec.~\ref{sec:core_framework}, before giving an overview of different toolkit building blocks in Sec.~\ref{sec:overview}. Then, we detail exposed degrees of freedom in Sec.~\ref{sec:dof} and explain the easy-to-implement interfaces that allow to fast adaption of the framework to new domains in Sec.~\ref{sec:adaptation}.

\subsection{Core Framework}
\label{sec:core_framework}
The \name{} toolkit is built upon the framework of Spatial Reasoning Models (SRMs)~\cite{wewer25srm}. Given a set of variables $\{x_1,...,x_n\}$, SRMs define \emph{reasoning} as an iterative denoising process over the set of variables:
\begin{equation}
\label{eq:reasoning}
    \hat{x}^{t_1}_1,...,\hat{x}^{t_n}_n \sim q(x^{t_1}_1,...,x^{t_n}_n \mid x^{t^\prime_1}_1,...,x^{t^\prime_n}_n) \textnormal{,}
\end{equation}
where $t_i$ encode individual noise levels for each variable. Depending on the task at hand, variables can represent different types of data, e.g. image patches, whole images of a sequence, or other entities, and can contain positional encodings to locate them in an arbitrary space.
For a denoising process of $d$ steps, a matrix $\mathbf{T}\in \mathbb{R}^{n\times d}$, containing the noise levels $t_i$ for all $n$ variables and all $d$ steps, fully specifies the reasoning process during inference. One step of the process can be carried out by predicting scores or flows, according to typical generative denoising formulations, such as DDPM~\cite{ddpm}, DDIM~\cite{ddim}, or Rectified Flows~\cite{liu2022rectifiedflow}.

\subsection{Overview}
\label{sec:overview}

Fig.~\ref{fig:pipeline} shows an overview of the full toolkit, where exchangeable and customizable building blocks are shown in yellow.  The \texttt{VariableMapper} can transform data from different domains into the \texttt{Variable} format, which is then unified for the rest of the pipeline. Training and inference routines work on top of this format. During training, noise is added to ground truth variables according to the chosen noise level sampling algorithm, before fed into the denoising model, which is trained to denoise them. During inference, variables can be (partially) initialized with random noise and denoised according to a defined schedule.

\subsection{Individual Degrees of Freedom}
\label{sec:dof}
The \name{} framework exposes a wide range of degrees of freedom to facilitate exploration and further research. We discuss them individually in the following.

\vspace{-0.3cm}
\paragraph{Denoising Paradigms.} We support original diffusion with discrete steps, implementing DDPM~\cite{ddpm} and DDIM~\cite{ddim}, diffusion with continuous steps, mixing diffusion and flow matching, cosine (variance preserving) flows~\cite{albergo2023stochastic, iddpm} and Rectified Flows~\cite{liu2022rectifiedflow}.

\vspace{-0.3cm}
\paragraph{Parameterizations.} \name{} supports a wide range of parameterizations, including $\epsilon$ prediction (noise), $x_0$ prediction (clean data), $u_t$ prediction (direction in flow models), $v$ prediction (direction vector in diffusion).

\vspace{-0.3cm}
\paragraph{Training $\mathbf{t}$-sampling.} We support different $\mathbf{t}$-samplers for training with sets of variables, such as the independent uniform sampling strategy~\cite{chen2024diffusionforcingnexttokenprediction}, or Uniform-$\bar{t}$ sampling~\cite{wewer25srm}. It is straightforward to implement additional sampling strategies. All $\mathbf{t}$-samplers for variable sets can be additionally combined with tailored scalar noise level samplers like the logit-normal distribution for Rectified Flows~\cite{sd3}.

\vspace{-0.3cm}
\paragraph{Architectures.} The framework makes it easy to exchange the neural architecture, which predicts the noise. Currently, we support DiT~\cite{dit}, LightningDiT~\cite{lightningdit}, UNet~\cite{ldm}, MAR~\cite{li2024autoregressive}, xAR~\cite{ren2025nexttokennextxpredictionautoregressive}, and U-ViT-Pose~\cite{song2025historyguidedvideodiffusion}. We support loading the checkpoints from the original works.

\vspace{-0.3cm}
\paragraph{Inference Schedules.} We support all the inference schedules from existing works, such as sequentialized sampling, with variable blend between autoregressive and parallel generation (overlap), and with predicted, manually-defined or random order~\cite{wewer25srm}, as well as next $k$ variable prediction~\cite{ren2025nexttokennextxpredictionautoregressive}. 

\vspace{-0.3cm}
\paragraph{Dependency Graph Injection.} We provide functionality to inject domain-specific knowledge by providing dependency structure between variables in form of graphs. Those can be exploited in choosing the order of inference.

\vspace{-0.3cm}
\paragraph{Uncertainty Prediction.} All models can easily parameterized to also predict uncertainty, allowing for uncertainty-based ordering of generation~\cite{wewer25srm}.

\vspace{-0.3cm}
\paragraph{Learned Variance.} By implementing a unified interface for different denoising paradigms, we support improvements for diffusion modes like the learning of the variance in the generative process~\cite{iddpm} also in combination with flow formulations like Rectified Flows~\cite{liu2022rectifiedflow}.

\vspace{-0.3cm}
\paragraph{Latent Denoising.} \name{} supports reasoning and generation in latent spaces, by including typical image autoencoders like SD-VAE~\cite{ldm}, VAVAE~\cite{lightningdit}, and DC-AE~\cite{dcae}.

\vspace{-0.3cm}
\paragraph{Modular Losses.} We support different losses including standard MSE for noise prediction, VLB~\cite{iddpm} for learning the variance of the reverse process, and cosine similarity for additional supervision of the velocity direction with flow models~\cite{yao2024fasterdit}. It is easy to add additional losses, e.g., other losses for uncertainty predictions besides NLL~\cite{wewer25srm}.

\subsection{Adapting to New Domains}
\label{sec:adaptation}
A main goal of \name{} is to make it easy to adopt SRMs to new data domains. We achieve this by providing two interface classes that need to be customized to support a new modality: the \texttt{VariableMapper} and the \texttt{Tokenizer}.

In the \texttt{VariableMapper}, the user needs to define how a data example should be partitioned into variables, atomic elements that maintain the same noise level. For latent space diffusion, an autoencoder can be defined here that pre-processes the data before partitioning.

The \texttt{Tokenizer} allows to transform the variables-format data to the arbitrary input format of the trained denoiser. Architectures such as DiT are domain agnostic and just require the definition of the token positions for positional encodings, e.g., sinusoidal,  RoPE, etc., depending on the generation task. 

In addition to the two mentioned interfaces, the user can implement custom visualization and metrics in \texttt{Evaluation} classes.
Thanks to the underlying variable format, the rest of the framework remains domain agnostic.

\section{Application Examples}
In this section, we provide a few application examples to showcase the generality of \name{}. We show examples for reasoning over image-based MNIST Sudoku~\cite{wewer25srm}, auto-regressive image generation~\cite{ren2025nexttokennextxpredictionautoregressive}, and auto-regressive, overlapping video generation~\cite{chen2024diffusionforcingnexttokenprediction, song2025historyguidedvideodiffusion}.

\label{sec:experiments}
\begin{figure*}[t!]
\centering
\begin{subfigure}[h]{\textwidth}
\centering
\includegraphics[width=0.85\textwidth]{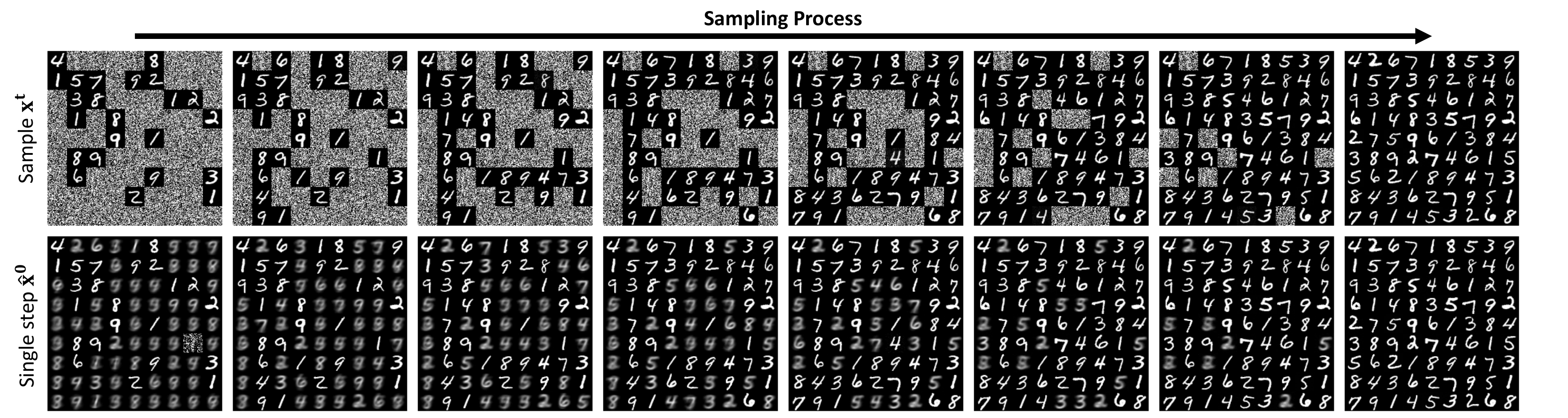}
\vspace{-0.2cm}
\caption{Visual Reasoning: Solving Visual Sudoku}
\label{fig:sudoku-gen}
\end{subfigure}
\begin{subfigure}[h]{\textwidth}
\centering
\includegraphics[width=\textwidth]{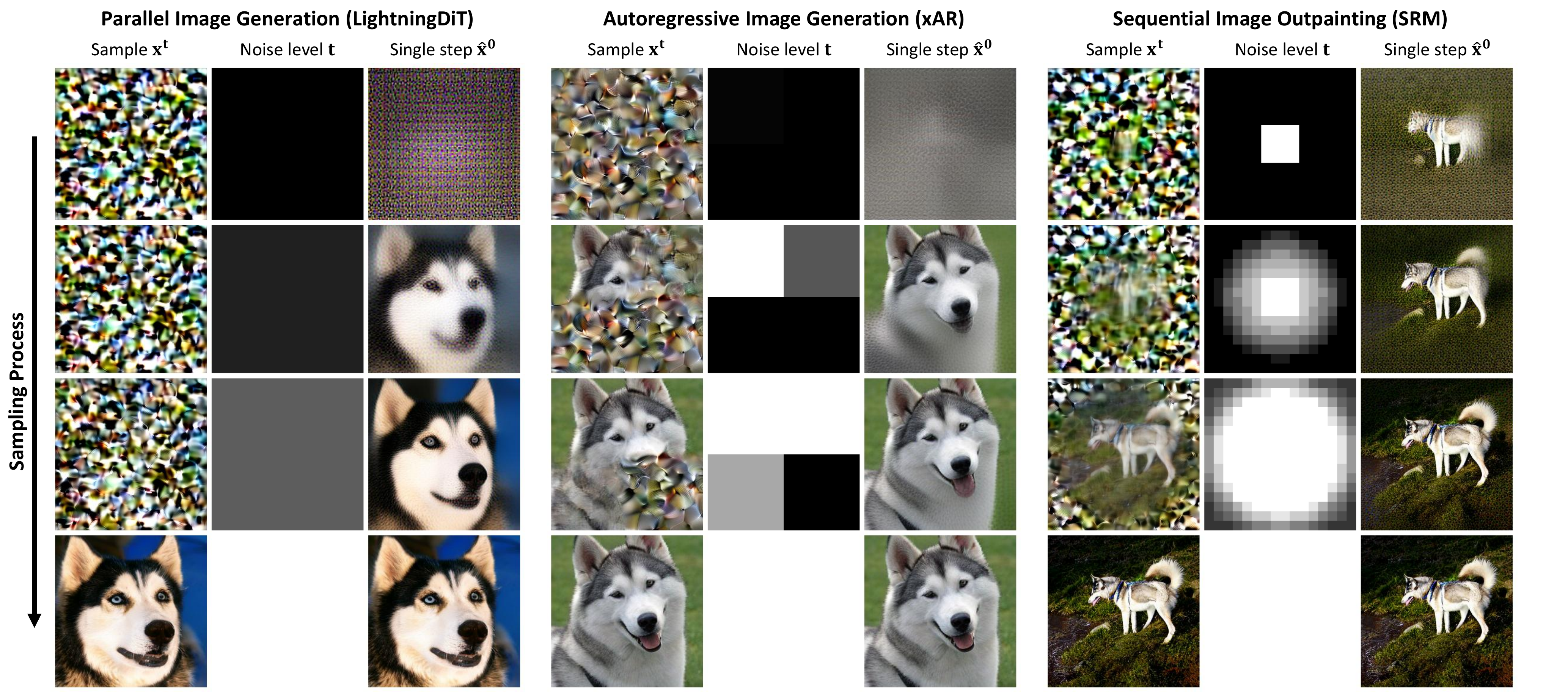}
\vspace{-0.5cm}
\caption{Latent Image Generation and Editing}
\label{fig:image-gen}
\end{subfigure}
\begin{subfigure}[h]{\textwidth}
\centering
\includegraphics[width=0.85\textwidth]{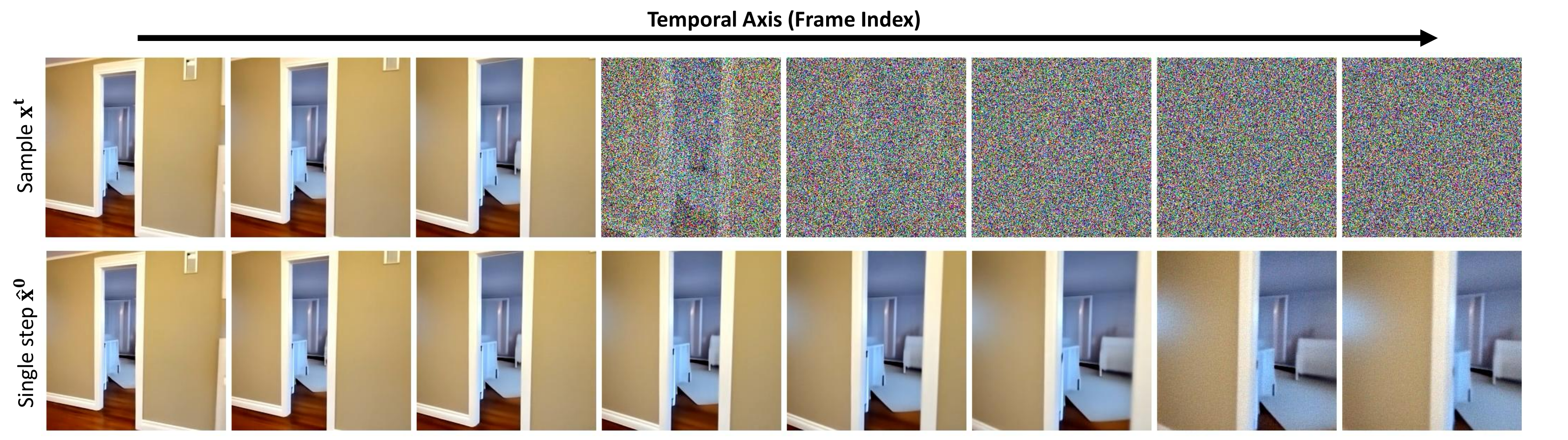}
\caption{Soft-Sequential Video Generation in Pixel Space}
\label{fig:video-gen}
\end{subfigure}

\caption{\textbf{Application Examples}. \name{} supports \textbf{(a)} visual reasoning benchmarks like visual Sudoku~\cite{wewer25srm}, \textbf{(b)} various image generation and editing strategies including parallel denoising with LightningDiT~\cite{lightningdit}, autoregressive next-X prediction with xAR~\cite{ren2025nexttokennextxpredictionautoregressive}, and soft, certainty-based sequentialization with Spatial Reasoning Models~\cite{wewer25srm}, and \textbf{(c)} auto-regressive, long horizon video generation, including history-guidance ~\cite{chen2024diffusionforcingnexttokenprediction, song2025historyguidedvideodiffusion}.
\vspace{-0.4cm}
}
\label{fig:pipeline}
\end{figure*}

\vspace{-0.3cm}
\paragraph{Visual Reasoning Tasks.} SRMs~\cite{wewer25srm} introduced multiple visual reasoning benchmarks, where variables are image patches. In Fig.~\ref{fig:sudoku-gen} we show sequential solving of visual Sudoku, consisting of MNIST numbers. It is fully auto-regressive with an order based on predicted uncertainty. The more numbers on the board, the less ambiguous the remaining ones, which is visible in the  $\mathbf{\hat{x}^0}$ prediction.  

\vspace{-0.3cm}
\paragraph{Image Generation and Editing.}
Fig.~\ref{fig:image-gen} shows multiple examples for image generation (left and middle), and out-painting (right). \name{} supports a variety of sampling schedules, such as standard parallel generation with LightningDiT~\cite{lightningdit}, next-$k$ variable generation of xAR~\cite{ren2025nexttokennextxpredictionautoregressive}, or manually defined schedules from SRMs
\cite{wewer25srm}. The SRM example (right) shows a locality-based order, painting outwards from existing variables. All shown models are latent diffusion models and generate in the latent space of a VAE.

\vspace{-0.3cm}
\paragraph{Soft-sequential Video Generation.}
We allow to perform soft-sequential video generation with a U-ViT model~\cite{song2025historyguidedvideodiffusion}, where each video frame is represented as one variable. Fig.~\ref{fig:video-gen} illustrates a moment during inference. While the first three frames are already fully denoised, the others are partially or fully noisy. However, the information from the already denoised frames and the camera pose conditioning is sufficient conditioning for the model to provide a good single-step $\mathbf{\hat{x}^0}$ prediction for fully noisy frames.

\section{Conclusion}
\label{sec:conclusion}
Denoising models have proven to be powerful tools for generative tasks, and recent developments have extended their utility to reasoning over multiple variables with distinct noise levels. By offering a clean, modular interface for defining variable mappers, training and inference schedules, together with access to a vast lineup of denoising architectures, \name{} aims to broaden the applicability of multi-noise-level generative models beyond traditional domains. We hope that \name{} will become a useful tool for researchers who want to take a deeper dive into probabilistic reasoning with structured generative models.

\section*{Acknowledgements}
This project was partially funded by the Saarland/Intel Joint Program on the Future of Graphics and Media. We thank Philipp Schröppel for insightful discussions regarding the design choices of the software architecture.

\bibliography{main}
\bibliographystyle{icml2025}

\end{document}